\documentclass[runningheads]{llncs}
\usepackage[T1]{fontenc}
\usepackage{amssymb}
\usepackage{utfsym}
\usepackage{graphicx}%
\usepackage{multirow}%
\usepackage{amsmath,amssymb,amsfonts}%
\usepackage{mathrsfs}%
\usepackage[title]{appendix}%
\usepackage{xcolor}%
\usepackage{textcomp}%
\usepackage{manyfoot}%
\usepackage{booktabs}%
\usepackage{algorithm}%
\usepackage{algorithmic}%
\usepackage{listings}%
\usepackage[utf8]{inputenc}
\usepackage{dutchcal}
\usepackage{tabularx} 
\usepackage{array}
\usepackage{makecell}
\usepackage{enumitem}
\usepackage[misc]{ifsym}
\graphicspath{ {image/} }

\begin{document}
\title{FHGE: A Fast Heterogeneous Graph Embedding with Ad-hoc Meta-paths}
%
%
\author{Xuqi Mao\inst{1,3} \and
Zhenying He\inst{1,3} \and
X. Sean Wang\inst{1,2,3}\textsuperscript{(\Letter)}}
\institute{School of Computer Science, Fudan University, Shanghai, China \and
School of Software, Fudan University, Shanghai, China \and
Shanghai Key Laboratory of Data Science, Shanghai, China\\
\email{\{xqmao17,zhenying,xywangCS\}@fudan.edu.cn}}
\maketitle              
\begin{abstract}
Graph neural networks (GNNs) have emerged as the state of the art for a variety of graph-related tasks and have been widely used in Heterogeneous Graphs (HetGs), where meta-paths help encode specific semantics between various node types. Despite the revolutionary representation capabilities of existing heterogeneous GNNs (HGNNs) due to their focus on improving the effectiveness of heterogeneity capturing, the huge training costs hinder their practical deployment in real-world scenarios that frequently require handling ad-hoc queries with user-defined meta-paths. To address this, we propose FHGE, a Fast Heterogeneous Graph Embedding designed for efficient, retraining-free generation of meta-path-guided graph embeddings. The key design of the proposed framework is two-fold: segmentation and reconstruction modules. It employs Meta-Path Units (MPUs) to segment the graph into local and global components, enabling swift integration of node embeddings from relevant MPUs during reconstruction and allowing quick adaptation to specific meta-paths. In addition, a dual attention mechanism is applied to enhance semantics capturing. Extensive experiments across diverse datasets demonstrate the effectiveness and efficiency of FHGE in generating meta-path-guided graph embeddings and downstream tasks, such as link prediction and node classification, highlighting its significant advantages for real-time graph analysis in ad-hoc queries.

\keywords{heterogeneous graph \and graph neural network \and ad-hoc meta-path.}
\end{abstract}
\section{Introduction}
\begin{figure}
\centering
\includegraphics[width=\textwidth]{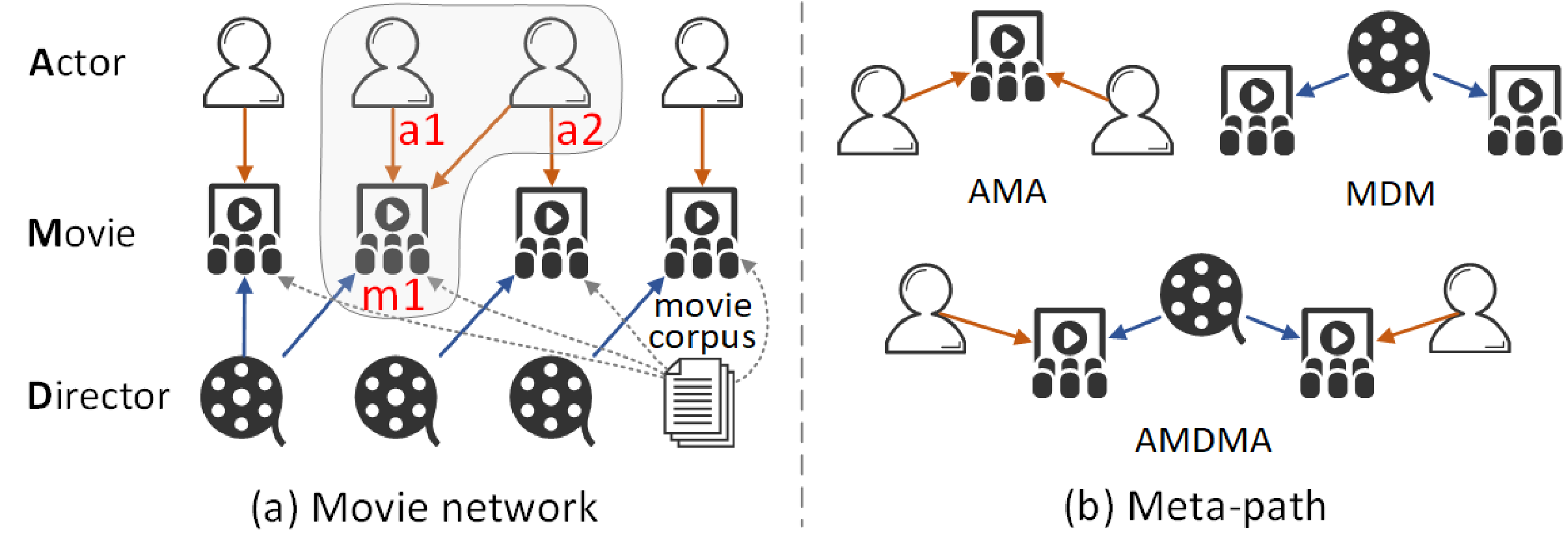}
\caption{Example of HetG. (a) depicts a movie network with heterogeneous nodes and relations, while (b) outlines its meta-paths. For instance, the path ``a1-m1-a2'' in (a) indicates that actor a1 and a2 co-starred in the movie m1, represented as the meta-path``AMA'' in (b).} \label{fig:example_of_hetg}
\end{figure}

Numerous real-world datasets can be modeled as Heterogeneous Graphs (HetGs), with diverse objects and the relations represented as various types of nodes and edges, respectively. Examples expand across various domains such as social networks \cite{gamba2024exit,muppasani2024expressive,avery2024effect}, citation networks \cite{jin2023heterformer,jia2023enhancing,mao2024hetfs}, traffic networks \cite{jin2023transferable}, recommendation systems \cite{agrawal2024no,liu2023generative,yang2024fine}, natural language processing \cite{Christmann23Explainable,MaYLMC24HetGPT,WU23Heterogeneous}, knowledge graphs \cite{zang2023commonsense,liu2023knowledge}, and biology information networks \cite{zhong2023knowledge}. Fig.~\ref{fig:example_of_hetg}(a) presents a movie graph example of a HetG, illustrating actors, movies, directors, and the interconnected relationships between them. The diversity of types and the richness of properties within HetGs give rise to a multitude of varied and intricate semantics and have led to a significant amount of related work, particularly in the domains of traditional graph representation learning and heterogeneous graph neural networks (HGNNs) \cite{Yu23KGTrust,Shi2022hgnn,yang23HGNAS}. 

The complex semantics of heterogeneous graphs often lead users to express preferences for specific relations. These preferences can be explicitly specified using meta-paths, i.e., sequences of node types and edge types defining composite relations between objects with examples in Fig.~\ref{fig:example_of_hetg}(b). Given their specificity, users often prefer node embeddings tailored to particular semantics. For example, consider ``Terminator 2'', a fictional action film directed by James Cameron and starring Arnold Schwarzenegger and Linda Hamilton. We analyzed the top 5 movies similar to it under different semantics. Table~\ref{tab:movies_under_meta-paths} showcases the results under distinct semantics: (a) movies with similar actors, using the meta-path ``MAM'', (b) movies with similar directors, using the meta-path ``MDM'', and (c) movies sharing all available similar information under ``meta-path-free'' semantics. The diverse outcomes for each scenario underscore the need to process ad-hoc queries with user-defined meta-paths.

\begin{table}[htbp]
\centering
\caption{Top 5 similar movies for ``Terminator 2'' under three different scenarios using FHGE.}\label{tab:movies_under_meta-paths}
\begin{tabular}{clll}
\toprule
Rank  & meta-path: MAM   & meta-path: MDM   & meta-path-free \\
\midrule
1     & Terminator & True Lies & Terminator \\
2     & True Lies & Aliens & Terminator 3 \\
3     & Total Recall & The Abyss & True Lies \\
4     & End of Days & Titanic & The Abyss \\
5     & Jingle All the Way & Ghosts of the Abyss & Titanic \\
\bottomrule
\end{tabular}%
\end{table}



Over the past decade, significant progress has been made in mining information from graphs from traditional representation learning approaches \cite{perozzi2014deepwalk,grover2016node2vec,dong2017metapath2vec} to methods utilizing deep neural networks, including GNNs \cite{fan2019metapath,yan2021relation,zhang23pagelink,zhu23AutoAC,SHAN24KPI-HGNN,MaYLMC24HetGPT} and GCNs \cite{kipf2016semi,liu2023rhgn}. Inspired by the Transformer \cite{vaswani2017attention}, GAT \cite{velickovic2017graph} integrates the attention to aggregate node-level information in homogeneous networks, while HAN \cite{wang2019heterogeneous} introduces a two-level attention mechanism for node and semantic information in heterogeneous networks. MAGNN \cite{fu2020magnn}, MHGNN \cite{liang2022meta} and R-HGNN \cite{yu23RHGNN} proposed meta-path-based models to learn meta-path-based node embeddings. HetGNN \cite{zhang2019heterogeneous} and MEGNN \cite{chang2022megnn} take a meta-path-free approach to consider both structural and content information for each node jointly. HGT \cite{hu2020heterogeneous} incorporates information from high-order neighbors of different types through messages passing across ``soft'' meta-paths. MHGCN \cite{fu2023multiplex} captures local and global information by modeling the multiplex structures with depth and breadth behavior pattern aggregation. SeHGNN \cite{Yang23Simple} simplifies structural information capture by precomputing neighbor aggregation and incorporating a transformer-based semantic fusion module. HAGNN \cite{zhu2023hagnn} integrates meta-path-based intra-type aggregation and meta-path-free inter-type aggregation to generate the final embeddings.



\begin{figure}[ht]
\centering
\includegraphics[width=\textwidth]{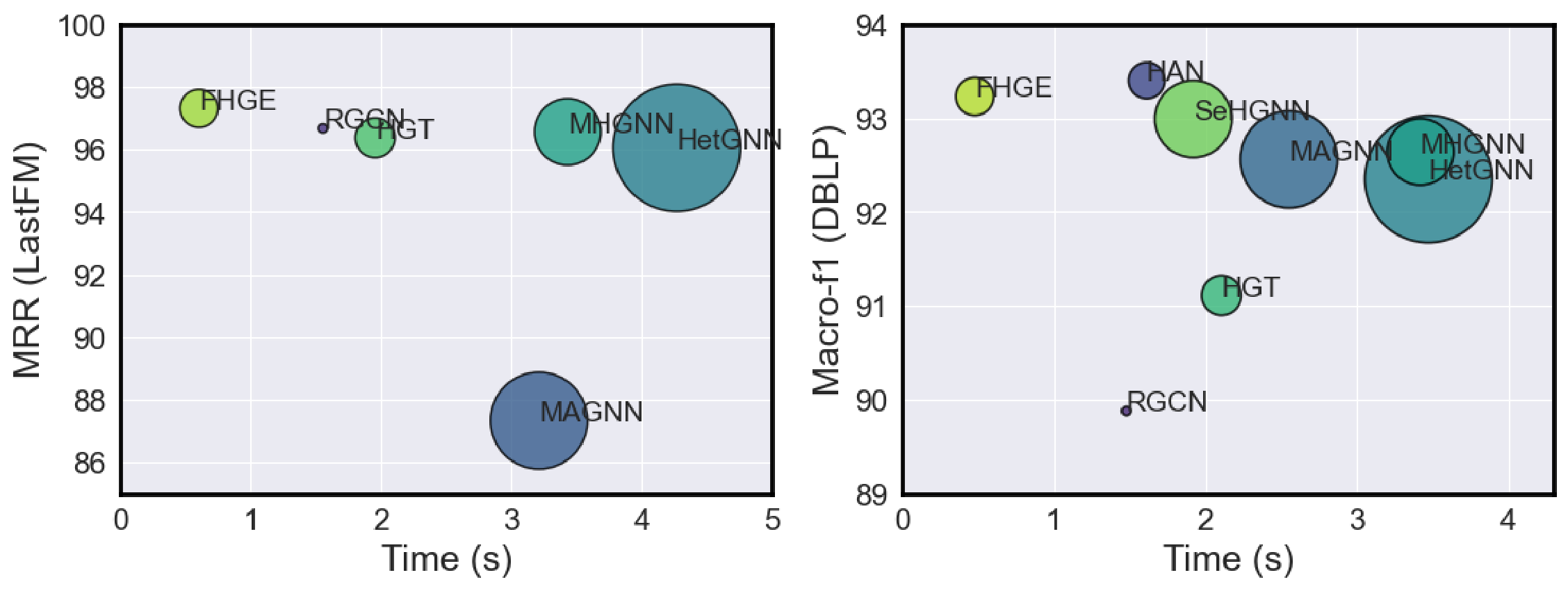}
\caption{Comparison of time and memory consumption for HGNNs on ``LastFM'' and ``DBLP''. The area of the circles indicates the memory consumption of each model, while the time costs are recorded on a log scale.} \label{fig:metric_time_memory}
\end{figure}

While existing methods have shown effectiveness in predefined meta-path and meta-path-free scenarios, they encounter significant challenges when generating meta-path-guided heterogeneous graph embeddings: (a) Meta-path-based HGNNs \cite{fu2020magnn,wang2019heterogeneous} generate accurate graph embeddings only when pre-defined meta-paths align the user-given meta-paths. Otherwise, the resulting embeddings will be suboptimal, and retraining the embeddings can be both time-consuming and practically challenging. (b) Meta-path-free HGNNs \cite{zhang2019heterogeneous,hu2020heterogeneous,li2023metapath,schlichtkrull2018modeling} either treat all meta-paths uniformly or assign weights to them using mechanisms like attention. However, these approaches may incorporate meta-paths irrelevant to the ad-hoc query, potentially leading to embeddings that do not align with the user-defined meta-paths. Fig.~\ref{fig:metric_time_memory} compares the time and memory overhead of various methods. This underscores the need for a method capable of meta-path-guided heterogeneous graph embedding.
 
To fill the gap outlined above, this paper proposed FHGE, a Fast Heterogeneous Graph Embedding designed for ad-hoc meta-paths. FHGE consists of two core modules: segmentation and reconstruction modules. It starts by segmenting graph information into local and global components using a Meta-Path Unit (MPU), then employs a random walk with a restart within the MPU to capture essential local structures. During reconstruction, FHGE quickly integrates node embeddings from relevant MPUs by reutilizing local embeddings, enabling swift adaptation to ad-hoc meta-paths. Specifically, it applies type-specific linear transformations to project heterogeneous content of nodes, potentially of unequal dimensions for different feature types, into the same latent vector space. A graph neural network architecture is then used to aggregate heterogeneous node information from both homogeneous and mixed-type nodes within the MPU. To enhance information capturing, a dual attention mechanism, one for intra-MPU and another for inter-MPU, is employed to mine local and global semantics. The aggregated features are further combined with attention based on ad-hoc meta-paths following two strategies, enabling FHGE to generate meta-path-guided graph embeddings that effectively leverage the structural, node, and semantic information.

To summarize, this work makes the following main contributions:
\begin{itemize}
\item To our best knowledge, FHGE is the first meta-path-guided heterogeneous network for fast heterogeneous graph embedding under ad-hoc meta-paths.
\item It introduces the MPU to separate graph information into local and global components. It allows the preservation of partial semantics on the heterogeneous graph, enabling its repetitive utilization.
\item A dual attention mechanism is applied to enhance graph information reconstruction, and two kinds of integration strategies are employed across various MPUs based on ad-hoc meta-paths.
\item Extensive experiments are conducted to demonstrate the accuracy and \\ prompt responsiveness of meta-path-guided graph embedding of FHGE.
\end{itemize}

\section{Preliminary}\label{sec:pre}

In this section, we give an overview of heterogeneous graphs.

\noindent\textbf{Definition 1} Heterogeneous Graph. A heterogeneous graph is a graph $G(V, E)$, where each node $v \in V$ is mapped to a specific node type $\theta(v) \in \mathcal{A}$ by $\theta: V \rightarrow \mathcal{A}$ and each edge $e \in E$ is mapped to a specific relation type $\kappa(e) \in \mathcal{R}$ by $\kappa: E \rightarrow \mathcal{R}$. In a heterogeneous graph, the number of distinct node types $|\mathcal{A}|$ is greater than 1, or the number of distinct relation types $|\mathcal{R}|$  is greater than 1. 
	
\noindent\textbf{Definition 2} Meta-path. A meta-path $\phi$ is a type sequence of nodes and edges defining a composite relation. It is typically represented by the notation of $A_1 \stackrel {R_1} {\longrightarrow} A_2 \stackrel {R_2} { \longrightarrow} \cdots \stackrel {R_l} {\longrightarrow} A_{l+1}$, where $A_1$ denotes the starting node type, $A_{l+1}$ denotes the ending node type, and each intermediate relation $R_i$ denotes a connection between node type $A_i$ and $A_{i+1}$. 

\section{Methodology}

\begin{figure*}[htbp]
\centering
\includegraphics[width=\textwidth]{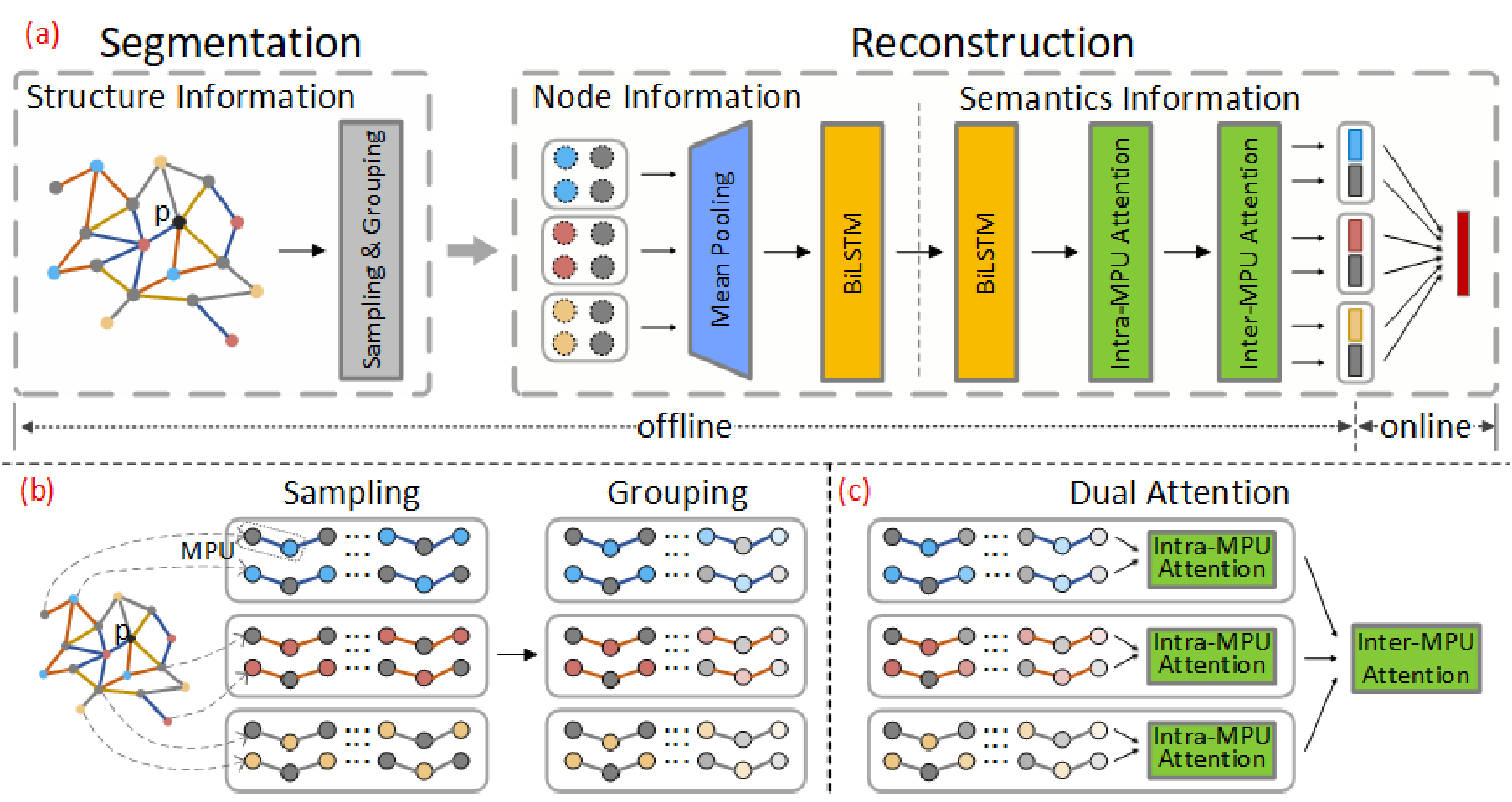}
\caption{(a) The overall architecture of FHGE: it first segments graph information into local and global components, samples and groups heterogeneous neighbors to balance the graph, and then reutilizes local information during the reconstruction process to facilitate meta-path-guided node embedding; (b) sampling and grouping process; (c) dual attention mechanism.} \label{fig:architecture}
\end{figure*}

In this section, we describe FHGE, which is designed for meta-path-guided graph embedding. FHGE is composed of two modules: segmentation and reconstruction module. Fig.~\ref{fig:architecture} illustrates the architecture of FHGE.

\subsection{Segmentation Module}

\subsubsection{Structure Information Extaction}

To efficiently generate meta-path-guided node embeddings, FHGE introduces a Meta-Path Unit (MPU) to separate graph information into local and global components. By preserving partial information across the entire heterogeneous graph, the MPU ensures repetitive utilization of node embedding within it, thereby enhancing the overall efficiency of the model. Next, we give the definition of MPU.

\noindent\textbf{Definition 3} Meta-path unit (MPU). An MPU $\psi$ is defined as the minimal segmentable unit of meta-path $\phi$, with a length of 1, represented as $\psi = (A_1A_2)$, $A_1$, $A_2 \in \mathcal{A}$ in the HetG. It serves as the smallest unit to compose a meta-path, capturing the fundamental relationships between two node types connected by an edge.



Before discussing the heterogeneous aggregation process, we first address the issue of unequal data distribution, which can negatively impact the embedding of ``hub'' nodes and inadequately represent ``cold-start'' nodes. To mitigate this, FHGE begins with MPU-based sampling using random walk with restart (RWR) in two steps: (a) Sampling Fixed-Length RWR. Beginning with a node $a \in V^\psi$, the random walker either moves to an adjacent node or returns to the starting node with probability $p$ until a fixed number of nodes for each type are collected. (b) Grouping Neighbors by Type. For each node type $A \in \mathcal{A}$, the top $k_A$ neighbors most frequently encountered during the RWR in MPU $\psi$ are selected as $A$-type neighbors, based on the intuition that frequently co-occurring nodes are more similar. This process enables FHGE to effectively capture and utilize the rich local information around each node, enhancing the quality of node embeddings. 

\subsection{Reconstruction Module}

\subsubsection{Node Information Transformation}
In heterogeneous graphs with node attributes, feature vectors for different content types of a node $a$ often have unequal dimensions and belong to distinct feature spaces. As a result, managing feature vectors of diverse dimensions within a unified framework presents significant challenges. To address this, a two-step transformation is used to capture node information.

\paragraph{Type-Specific Transformation} 

For each content type of a node, features are projected into a unified feature space via a type-specific neural network $f$. For the $n$-th content of node $a$, we have:
\begin{equation}
H_{an}=f_n\big(C_{an}\big)
\end{equation}
where $C_{an}$ is the original feature vector of $n$-th content for node $a$ and $H_{an}$ is the projected latent vector, $f_n$ is the type-specific transformation function, which can be pre-trained using different techniques for various content types. The node information of node $a$ can then be represented as:
\begin{equation}
H_a=\left\{H_{an}, n \in |C_a|\right\}
\end{equation}
where $H_a$ is the collection of latent feature vectors of node $a$, and $|C_a|$ denotes the amount of content feature of node $a$. 


\paragraph{Node Information Aggregation} 

After transforming the content for each node, all node features are standardized to the same dimension. We utilize Bidirectional LSTM, as described by \cite{hamilton2017inductive}, to aggregate the diverse set of unordered features of the node. The feature embedding process is as follows:
\begin{equation}
\hat{H}_a = \frac{\sum_{n\in H_a} \left[\overrightarrow{LSTM}\left (H_{an}\right) \bigoplus \overleftarrow{LSTM}\left (H_{an}\right)\right]}{|H_a|}
\end{equation}
where $\bigoplus$ denotes concatenation. This method enables the integration of features from neighbors of the same type, ensuring comprehensive representation in the embeddings. This operation aligns the projected features of all nodes to a uniform dimension, facilitating the subsequent processing of nodes of arbitrary types. 



\subsubsection{Semantics Information Aggregation}

To efficiently aggregate heterogeneous neighbor embeddings for each node, we developed an MPU-based aggregation network with two main components: (1) intra-MPU aggregation and (2) inter-MPU aggregation.

\paragraph{Intra-MPU Aggregation}

Given an MPU $\psi$, the intra-MPU aggregation captures local information inherent in the target node, its MPU-based neighbors, and the contextual relationships in between. The aggregation of fixed-size neighbor sets corresponding to each node type for each node $a$ is achieved using LSTM, formulated as follows:
\begin{equation}
Q_a = \frac{\sum_{b \in N_a} \left[\overrightarrow{LSTM}\left(\hat{H}_b\right) \bigoplus \overleftarrow{LSTM}\left(\hat{H}_b\right)\right]}{|N_a|}
\end{equation}
where $N_a$ denotes the sampled neighbor set of $a$, $b$ is one of the neighbors of $a$ within MPU $\psi$. Subsequently, we apply a graph attention layer \cite{velickovic2017graph} to compute a weighted sum of the nodes in each MPU $\psi$ related to the target node $a$. The attention mechanism is structured as follows:
\begin{equation}
\begin{aligned}
&\alpha_{a,b} = \frac{exp{\left(LeakyReLU\left(u^T[{Q}_a \bigoplus {Q}_b]\right)\right)}}{\sum_{s\in N_a}exp{\left(LeakyReLU\left(u^T[{Q}_a \bigoplus {Q}_s]\right)\right)}}\\
&Zi_a = \sum\nolimits_{b \in N_a}\alpha_{a,b} \odot {Q}_b
\end{aligned}
\end{equation}
where $\alpha_{a,b}$ is the normalized weight of node $b$ to node $a$, $u^T$ denotes the node-level attention vector, and $LeakyReLU$ refers to the leaky version of a Rectified Linear Unit, $Zi_a$ is the node embedding of $a$ within MPU $\psi$, $\odot$ denotes Hadamard product.

\begin{figure}[htbp]
    \centering
        \includegraphics[width=0.65\textwidth]{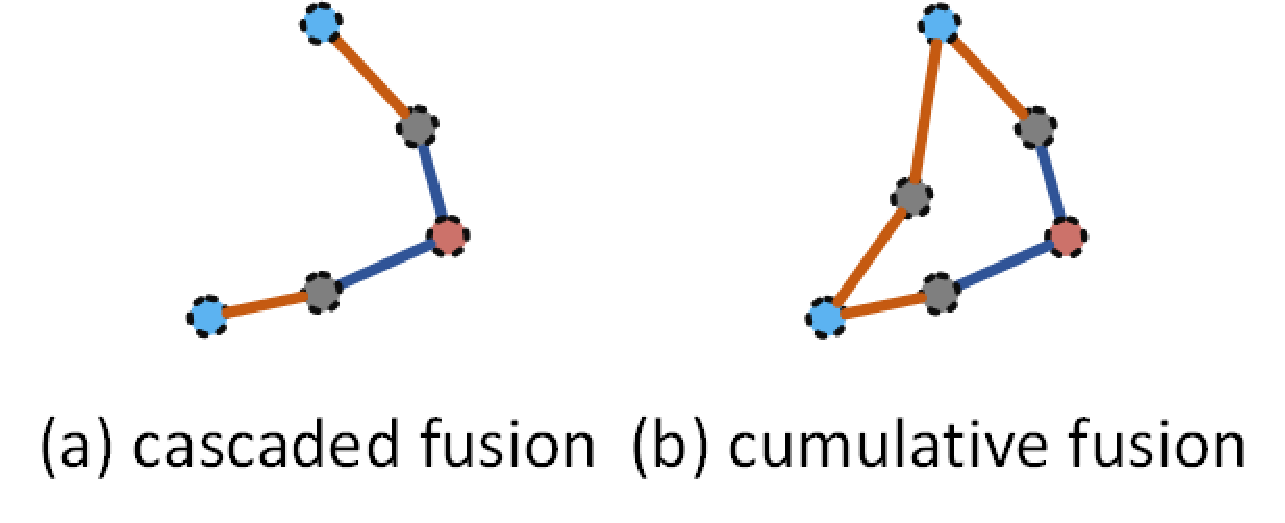}
    \caption{Two types of embedding integration based on user-defined meta-paths.} \label{fig:integration}
\end{figure}

\paragraph{Inter-MPU Aggregation}
After aggregating the local information within each MPU, the next step is inter-MPU aggregation, which involves combining the semantic information across all MPUs. First, we employ an attention mechanism to assign varying weights $\beta^{\psi}$ to different MPUs:
\begin{equation}
\begin{aligned}
&\beta^{\psi} = \frac{exp\left(LeakyReLU\left(q^T \odot Zi_a\right)\right)}{\sum\nolimits_{i\in|\psi|}exp\left(LeakyReLU\left(q^T \odot Zi_{ai}\right)\right)}\\
\end{aligned}
\end{equation}
where $\beta_{\psi}$ is the normalized weight of MPU $\psi$ over all MPUs, $q^T$ represents the parameterized attention vector. Clearly, a higher value of $\beta_a$ indicates a greater importance of MPU $\psi$. Subsequently, with node embeddings on the MPU, FHGE can generate graph embeddings for any user-defined meta-path. The integration of node embeddings can be divided into two scenarios: cascaded integration and cumulative integration, as illustrated in fig.~\ref{fig:integration}.

Cascaded Integration. In a movie graph context, for example, exploring actors starring in movies directed by the same director requires integrating ``AM'' and ``MD'' MPUs for ``deeper'' interactions. We implement cascaded integration to capture this specific semantics for meta-path $\phi$ = ($A_1A_2 \cdots A_{l\text{-}1} \\ A_lA_{{\text{-}}l+1} \cdots A_{\text{-}2}A_{\text{-}1}$):
\begin{equation}
\begin{aligned}
&Zw_a^{\psi} = \left\|\beta^{\psi}\right\|_1 \odot Zi_a\\
&Z^{\phi}_a = Zw^{1}_{a} \odot Zw^{2}_{a} \cdots \odot Zw^{{l\text{-}1}}_{a} \\
\end{aligned}
\end{equation}
where ${l\text{-}1}$ indicates the length of the meta-path $\phi$, representing the depth of interactions.

Cumulative integration. Here, we accumulate node embeddings from different meta-paths. For instance, one might be interested in actors who have collaborated as well as actors starring in movies directed by the same director, involving two meta-paths, namely ``AMA'' and ``AMDMA''. For such cases, we use cumulative integration to integrate ``broader'' semantics:
\begin{equation}
Z^{\phi}_a = \frac{\sum_{j\in|\phi|}Zw^{{j1}}_{a_{j1}} \odot Zw^{{j2}}_{a_{j2}} \cdots \odot Zw^{{jl}}_{a_{j{(l\text{-}1)}}}}{|\phi|}
\end{equation}
where $|\phi|$ is the number of meta-paths the user is interested in, and $a_{j{(l\text{-}1)}}$ represents the node at depth $l\text{-}1$ in the $j$-th meta-path. With these techniques, FHGE can generate aligned node embeddings for any user-defined meta-paths.


\subsection{Training}

We optimize the model weights by minimizing the cross-entropy loss function using negative sampling \cite{mikolov2013distributed}:
\begin{equation}
\mathcal{L} = \sum\limits_{\langle a, b, b^{\prime}\rangle \in \tau} \log \sigma \Big(Zi_a \cdot Zi_b\Big) - \log \sigma \Big(\text{-}Zi_a \cdot Zi_{b^{\prime}}\Big)
\end{equation}
where $\sigma(\cdot)$ represents the sigmoid function, $\tau$ denotes the set of triples $\langle a, b, b^{\prime}\rangle$ collected by walk sampling based on MPU $\psi$ in the heterogeneous graph. 


\section{Experiments} \label{sec:experiment}

In this section, we conduct extensive experiments to study the following research questions: \textbf{RQ1}: How does FHGE compare to SOTA methods in terms of effectiveness and efficiency for meta-path-guided graph embedding? \textbf{RQ2}: How does FHGE perform relative to SOTA methods across various downstream graph mining tasks such as link prediction and node classification? \textbf{RQ3}: What impact do different components, such as global and local semantics information, have on the performance of FHGE? \textbf{RQ4}: How do different hyper-parameters affect the performance of FHGE?

\subsection{Experimental Setup}

We describe the details of datasets and methods, how to set the ground truth and parameters, and how to assess the effectiveness of each method in our experiments.

\subsubsection{Datasets}

We use seven real-world datasets, i.e., Academic, IMDB, Amazon, LastFM, PubMed, DBLP, and ACM, which are widely used for benchmarking \cite{lv2021we}. The main statistics of the seven datasets are summarized in Table 3.

\begin{table}[htbp]
\centering
\caption{\centering Statistics of datasets.}\label{tab:statisticOfDatasets}
    \begin{tabular}{cccccccc}
    \toprule
    & Academic & IMDB & Amazon & LastFM & PubMed & DBLP & ACM \\
    \midrule
    Node & 72258 & 28751 & 10099 & 20612 & 63109 & 26128 & 10942 \\
    Edge & 296367 & 47355 & 148659 & 141521 & 244986 & 239566 & 547872 \\
    \bottomrule
    \end{tabular}%
\end{table}

\subsubsection{Baselines}

We compare FHGE against the following baselines: (1) SeHGNN \cite{Yang23Simple}: an HGNN adopts a light-weight mean aggregator to reduce complexity by removing overused neighbor attention and avoiding repeated neighbor aggregation. (2) HGT \cite{hu2020heterogeneous}: an HGNN that extends the transformer architecture to the graph-structured data, utilizing the self-attention mechanism to capture heterogeneous graph characteristics. (3) MHGNN \cite{li2023metapath}: a meta-path-based HGNN capturing both structural and content-related information from HetGs. (4) HetGNN \cite{zhang2019heterogeneous}: an HGNN using attention mechanisms to aggregate both structural and content information from heterogeneous neighbors. (5) MAGNN \cite{fu2020magnn}: an HGNN capturing information at various levels by incorporating node attributes, intermediate semantic nodes, and multiple meta-paths. (6) HAN \cite{wang2019heterogeneous}: an HGNN firstly employing a two-layer attention mechanism to capture weights at both the node and semantic levels in HetGs. (7) RGCN \cite{schlichtkrull2018modeling}: an extension of GCN for relational (multiple edge types) graphs that combines graph convolutions tailored to specific edge types.


\subsubsection{Evaluation Metrics}

To assess the effectiveness of FHGE, we employ two widely-used evaluation metrics, Recall@K and NDCG@K, focusing on the recommendation results from meta-path-guided node embeddings with $K$ set to 20 by default. To evaluate embedding efficiency, we track the CPU time required. To gauge the performance in downstream applications, we conduct meta-path-free experiments. Specifically, we use ``AUC'' and ``MRR'' for link prediction tasks and ``Micro-F1'' and ``Macro-F'' for node classification tasks.


\subsubsection{Implementation Details}

The whole method is implemented on PyTorch. We initialize the trainable parameters with Xavier initialization and optimize loos with Adam. As for hyper-parameters, we decide the important hyper-parameters by grid search and keep them the same in all datasets. For example, the size of embeddings is 128, the learning rate is 0.01. We configure the random walk parameters with a window size of 5, a walk length of 30, 10 walks per node, and 5 negative samples. Following the previous work, we also use early stopping to terminate training with a patient of 20 epochs, and the maximum training epochs is 200. For the benchmark methods, We utilize the code provided by a uniform benchmark \cite{lv2021we}. 




\subsection{Meta-Path-Guided Graph Embedding(RQ1)}

As mentioned earlier, various meta-paths convey varied semantic meanings, leading to diverse graph embedding that influences query results. We focus on evaluating the effectiveness and efficiency of the method for meta-path-guided graph embedding.

\subsubsection{Effectiveness}


Given that MHGNN generates distinct node embeddings for each meta-path and excels in downstream applications such as link prediction \ref{sec:app}, we validate our meta-path-guided embedding performance using the recommendation results of MHGNN as the benchmark. Table~\ref{tab:recommendation} shows that the FHGE model demonstrates superior performance in both datasets, confirming its effectiveness in feature extraction and reconstruction. MAGNN and SeHGNN also show strong performances in specific categories, especially in handling complex structural data. The performance of RGCN suggests that simple aggregation of graph information may not be enough for analysis.

\begin{table}[htbp]
\centering
\caption{\centering recommendation result using meta-path-guided node embeddings. r@20 refers to recall@20 and n@20 refers to ndcg@20. Vacant positions (``-'') are due to a lack of node embeddings based on that method. The best results are highlighted in bold.}\label{tab:recommendation}
\begin{tabular}{lcccccccc}
\toprule
      & \multicolumn{4}{c}{LastFM}    & \multicolumn{4}{c}{IMDB} \\
\midrule
      & \multicolumn{2}{c}{UAU} & \multicolumn{2}{c}{UATAU} & \multicolumn{2}{c}{MAM} & \multicolumn{2}{c}{MGM} \\
\midrule
      & \multicolumn{1}{l}{r@20} & \multicolumn{1}{l}{n@20} & \multicolumn{1}{l}{r@20} & \multicolumn{1}{l}{n@20} & \multicolumn{1}{l}{r@20} & \multicolumn{1}{l}{n@20} & \multicolumn{1}{l}{r@20} & \multicolumn{1}{l}{n@20} \\
\midrule
RGCN  & 12.41 & 9.39  & 7.97  & 6.09  & 63.58 & 54.17 & 86.21 & 74.63 \\
MAGNN & 13.66 & 12.05 & 9.81  & 9.32  & 65.44 & 56.35 & 87.73 & 73.21 \\
SeHGNN & \multicolumn{1}{l}{-} & \multicolumn{1}{l}{-} & \multicolumn{1}{l}{-} & \multicolumn{1}{l}{-} & 68.52 & 56.89 & \textbf{88.65} & \textbf{74.79} \\
FHGE  & \textbf{15.32} & \textbf{14.68} & \textbf{13.99} & \textbf{12.24} & \textbf{69.33} & \textbf{57.2}  & 88.19 & 73.24 \\
\bottomrule
\end{tabular}%
\end{table}%

\subsubsection{Efficiency}

\begin{figure}[htbp]
    \centering
        \includegraphics[width=\textwidth]{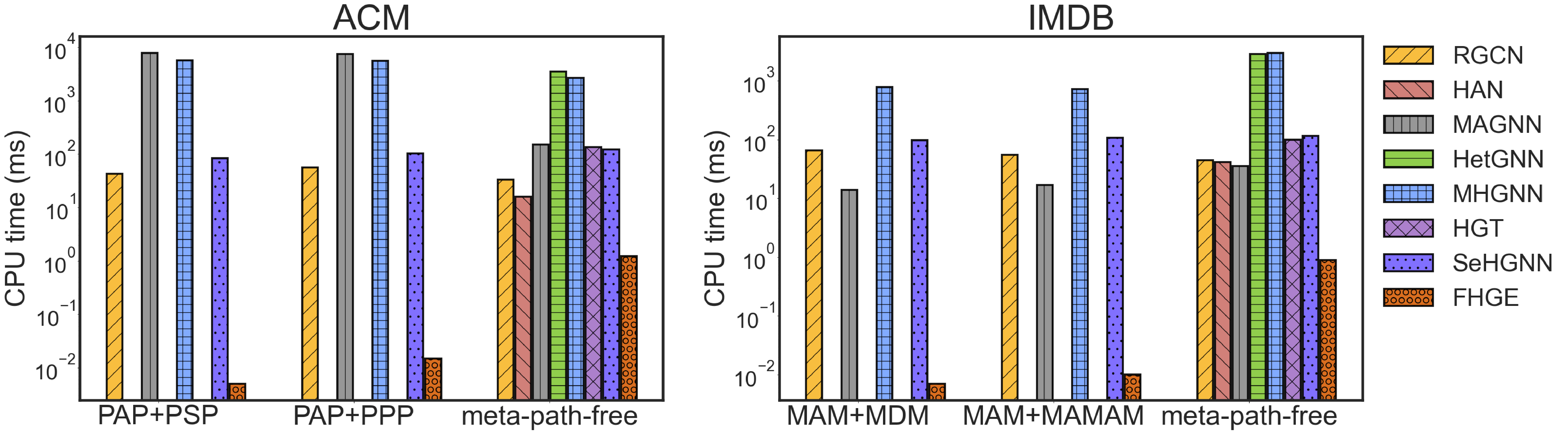}
    \caption{The time cost comparison on different meta-paths for the ``Academic'' and ``IMDB'' datasets. The time costs are presented on a log scale. (Blank bars indicate that the method is unable to generate meta-path-guided graph embeddings.)} \label{fig:time_cost_compare}
\end{figure}

To accommodate ad-hoc meta-paths defined by users, it's crucial that methods can quickly generate meta-path-guided graph embeddings. Fig.~\ref{fig:time_cost_compare} depicts the efficiency performance of these methods based on various scenarios. In the meta-path-free scenario on the ACM dataset, HetGNN and MHGNN take 3523.5s and 2693.7s, respectively, while RGCN, HAN, and HGT take 32.76s, 15.92s, and 135.57s, respectively. Conversely, FHGE achieved a time cost of just 1.2s, 26.6 times faster than the HAN on the ACM dataset and 39.8 times faster than the MAGNN method on the IMDB dataset. The comparison of the time cost of node embedding based on specific meta-paths gives a similar conclusion. This rapid performance is reaffirmed in Fig. ~\ref{fig:metric_time_memory}, which compares the time cost of node embedding for specific meta-paths across different datasets. The exceptional speed of FHGE in meta-path-guided node embedding makes it capable of user query with ad-hoc meta-paths.

\subsection{Applications (RQ2)}\label{sec:app}

To validate the practicality of HetFS, we compare it with SOTA methods on various applications.

\subsubsection{Link Prediction}\label{subsec:lp}
Table~\ref{tab:lp} presents the performance metrics of all models, with the superior results highlighted in bold. FHGE demonstrates competitive performance compared to SOTA methods. Despite not spending extensive time generating graph embedding online, FHGE still achieves strong results in link prediction. This effectiveness is due to its systematic integration of structural, node, and semantic information, which adeptly captures the complexities of graph data. Additionally, the dual attention mechanism employed by FHGE effectively discerns subtle semantics, preserving the intrinsic relationships between nodes.

\begin{table}[htbp]
\centering
\caption{\centering Link prediction task. Vacant positions (``-'') are due to a lack of similarities based on that method. Highlighted are the top \underline{\textbf{first}}, \textbf{second}, \underline{third}.}\label{tab:lp}
\begin{tabular}{ccccccc}
\toprule
      & \multicolumn{2}{c}{Amazon} & \multicolumn{2}{c}{LastFM} & \multicolumn{2}{c}{PubMed} \\
\midrule
      & AUC   & MRR   & AUC   & MRR   & AUC   & MRR \\
\midrule
RGCN  & 85.02 & 93.76 & 81.91 & \textbf{96.67} & 71.79 & \underline{\textbf{97.37}} \\
MAGNN & -     & -     & 81.06 & 87.33 & -     & - \\
HetGNN & \textbf{95.53} & \underline{\textbf{97.86}} & 87.73 & 96.05 & 89.45 & 94.21 \\
MHGNN & \textbf{95.53} & \textbf{97.85} & \textbf{87.93} & \underline{96.56} & \underline{89.89} & 94.57 \\
HGT   & 95.3  & 97.79 & \underline{\textbf{90.22}} & 96.37 & \underline{\textbf{90.43}} & \textbf{96.98} \\
FHGNN & \underline{\textbf{95.54}} & \textbf{97.85} & \textbf{89.26} & \underline{\textbf{97.31}} & \textbf{90.38} & \underline{96.65} \\
\bottomrule
\end{tabular}%
\end{table}%

\subsubsection{Node Classification}

Table~\ref{tab:nc} reports the Micro-F1 and Macro-F1 scores for each model, with the best results highlighted in bold. As we can see: (1) Most models demonstrate excellent performance in single-label classification, achieving high Macro-F1 and Micro-F1 scores (over 0.80). This is expected given the distinct nature of labeled nodes. (2) MAGNN and FHGE demonstrate high consistency and effectiveness across both datasets. This is because their architectures are carefully designed to capture information from different relations. Consequently, they are well-suited for diverse data scenarios.

\begin{table}[htbp]
\centering
\caption{\centering Node classification task. Highlighted are the top \underline{\textbf{first}}, \textbf{second}, \underline{third}.}\label{tab:nc}
\begin{tabular}{ccccc}
\toprule
      & \multicolumn{2}{c}{DBLP} & \multicolumn{2}{c}{ACM} \\
\midrule
      & Macro-F1 & Micro-F1 & Macro-F1 & Micro-F1 \\
\midrule
RGCN  & 89.88 & 90.49 & 83.14 & 82.91 \\
HAN   & \underline{\textbf{93.4}}  & \textbf{93.84} & 86.78 & 86.87 \\
MAGNN & \underline{92.56} & \underline{93.24} & \underline{\textbf{92.05}} & \underline{\textbf{91.97}} \\
HetGNN & 92.35 & 92.88 & 88.76 & 88.62 \\
MHGNN & \textbf{92.64} & 92.95 & \underline{90.83}  & \underline{90.77} \\
HGT   & 91.11 & 92.04 & 88.82 & 88.91 \\
FHGE & 92.05 & \underline{\textbf{93.97}} & \textbf{90.86}  & \textbf{90.79} \\
\bottomrule
\end{tabular}%
\end{table}


\subsection{Ablation Study (RQ3)}

FHGE introduces a dual attention mechanism to capture local and global semantics by intra-MPU and inter-MPU aggregation, respectively. To validate these attentions for FHGE, we conduct ablation studies to evaluate the performance of two model variants that eliminate the intra-MPU and inter-MPU attention, respectively. Table~\ref{tab:ablation_study} shows that both intra-MPU attention and inter-MPU attention consistently improve overall performance. However, intra-MPU attention contributes to a smaller improvement across various tasks, potentially due to the presence of excessive irrelevant information, particularly relational data, within the graph, which may degrade overall performance effectiveness.

\begin{table}[htbp]
\centering
\caption{\centering Ablation study for FHGE.}\label{tab:ablation_study}
\begin{tabular}{ccccccc}
\toprule
Task & \multicolumn{2}{c}{Link Prediction} & \multicolumn{2}{c}{Node Classification} \\
\midrule
Dataset & \multicolumn{2}{c}{LastFM} & \multicolumn{2}{c}{DBLP} \\
\midrule
Metric & AUC & MRR & Macro-F1 & Micro-F1 \\
\midrule
+intra-MPU attention & 85.88 & 94.49 & 83.14 & 84.91 \\
+inter-MPU attention & 88.56 & 94.24 & 86.78 & 88.87 \\
FHGE & 89.26  & 97.31 & 92.05 & 93.97 \\
\bottomrule
\end{tabular}%
\end{table}

\subsection{Hyper-parameter Sensitivity (RQ4)}

The hyper-parameters play important roles in FHGE. To provide in-depth knowledge of how they determine the generations of graph embedding, we investigate the sensitivity of different hyper-parameters in FHGE, including the amount of iterations and the embedding dimensions.

\subsubsection{Amount of Iterations}

\begin{figure}[htbp]
    \centering
        \includegraphics[width=\textwidth]{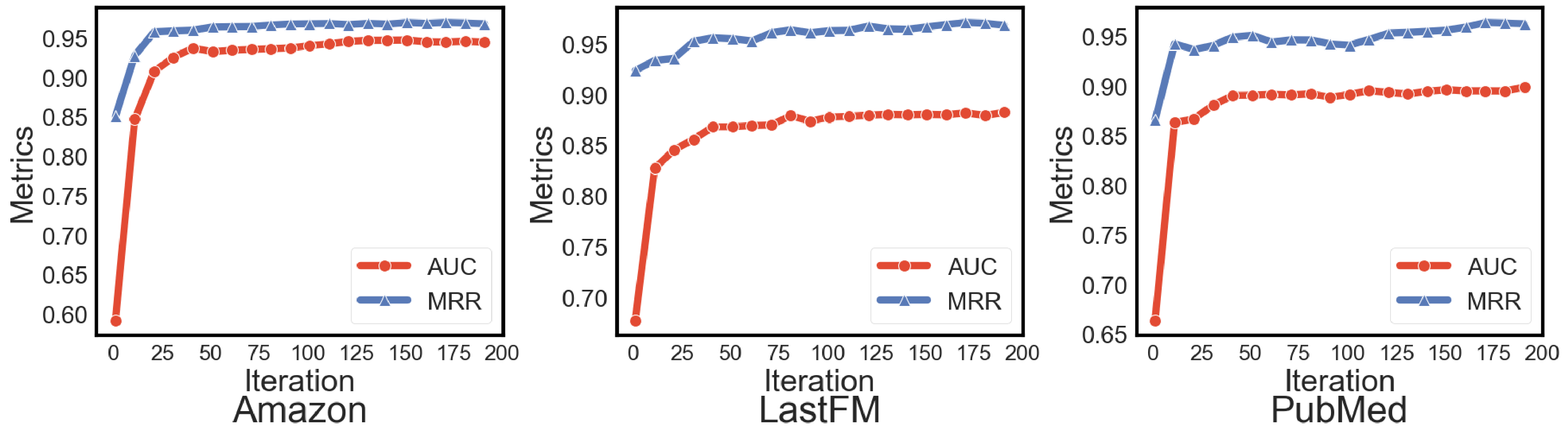}
    \caption{Impact of iteration amount on link prediction tasks.} \label{fig:iteration_lp}
\end{figure}

We analyze the convergence properties of our FHGE on various datasets. The results in Fig.~\ref{fig:iteration_lp} demonstrate that FHGE converges quickly and reaches stable performance after 50 iterations, with additional iterations not significantly enhancing its performance. This may be attributed to the increased perceptual field during the precomputation phase, which allows for more comprehensive information capture. The ability to quickly converge makes FHGE particularly well suited for generating ad hoc meta-path-guided node embeddings.



\subsubsection{Amount of Embedding Dimensions}

\begin{figure}[htbp]
    \centering
        \includegraphics[width=\textwidth]{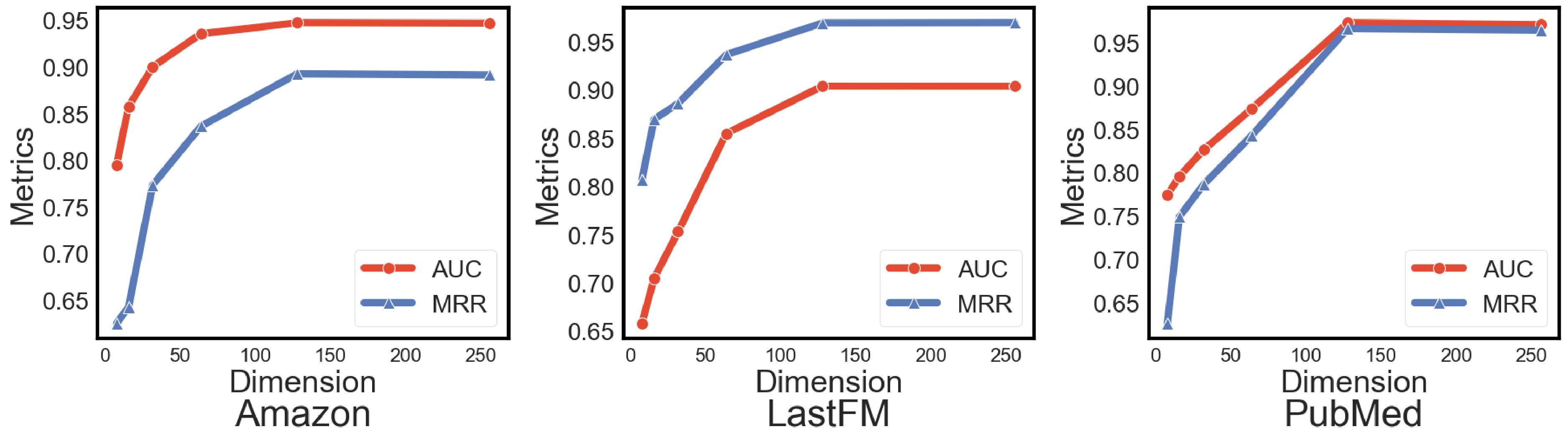}
    \caption{Impact of dimension amount on link prediction tasks.} \label{fig:dimension}
\end{figure}

We then test the effect of the embedding dimension. Fig.~\ref{fig:dimension} illustrates the performance of FHGE when altering the embedding dimension from the default setting 128. We can see that as the embedding dimension grows, the performance rises first, then holds on within a large range of embedding dimensions, and even drops when the embedding dimension is too large. This is because the more embedding dimensions, the better representations can be learned. However, a too-large dimension may introduce additional redundancies and cause overfitting to the model.

\section{Discussion and Conclusion}

In this study, we present FHGE, a fast heterogeneous graph embedding method designed for ad-hoc queries, which incorporates graph information segmentation and reconstruction. Specifically, FHGE employs MPU to segment graph information into local and global components, then integrates node embeddings from relevant MPUs during reconstruction, while reusing local information to quickly adapt to specific meta-paths. During reconstruction, a dual attention mechanism is employed to enhance semantics capturing. Extensive experiments confirm its effectiveness and efficiency in similarity search and various downstream applications. Future research could focus on enhancing multi-label node embedding, extending FHGE to temporal link prediction, and boosting performance using advanced language models.

\begin{credits}
\subsubsection{\ackname} This work was mainly supported by the National Natural Science Foundation of China (NSFC No. 61732004).

\end{credits}

%
%
%
%
\bibliographystyle{splncs04}
\bibliography{FHGE_D}

\begin{thebibliography}{10}
\providecommand{\url}[1]{\texttt{#1}}
\providecommand{\urlprefix}{URL }
\providecommand{\doi}[1]{https://doi.org/#1}

\bibitem{agrawal2024no}
Agrawal, N., Sirohi, A.K., Kumar, S., et~al.: No prejudice! fair federated graph neural networks for personalized recommendation. In: AAAI. vol.~38, pp. 10775--10783 (2024)

\bibitem{avery2024effect}
Avery, K., Houmansadr, A., Jensen, D.: The effect of alter ego accounts on a/b tests in social networks. In: WWW. pp. 565--568 (2024)

\bibitem{chang2022megnn}
Chang, Y., Chen, C., Hu, W., Zheng, Z., Zhou, X., Chen, S.: Megnn: Meta-path extracted graph neural network for heterogeneous graph representation learning. Knowledge-Based Systems  \textbf{235},  107611 (2022)

\bibitem{Christmann23Explainable}
Christmann, P., Saha~Roy, R., Weikum, G.: Explainable conversational question answering over heterogeneous sources via iterative graph neural networks. In: SIGIR. p. 643–653 (2023)

\bibitem{dong2017metapath2vec}
Dong, Y., Chawla, N.V., Swami, A.: metapath2vec: Scalable representation learning for heterogeneous networks. In: SIGKDD. pp. 135--144 (2017)

\bibitem{fan2019metapath}
Fan, S., Zhu, J., Han, X., Shi, C., Hu, L., Ma, B., Li, Y.: Metapath-guided heterogeneous graph neural network for intent recommendation. In: SIGKDD. pp. 2478--2486 (2019)

\bibitem{fu2023multiplex}
Fu, C., Zheng, G., Huang, C., Yu, Y., Dong, J.: Multiplex heterogeneous graph neural network with behavior pattern modeling. In: SIGKDD. pp. 482--494. {ACM} (2023)

\bibitem{fu2020magnn}
Fu, X., Zhang, J., Meng, Z., King, I.: Magnn: Metapath aggregated graph neural network for heterogeneous graph embedding. In: WWW. pp. 2331--2341 (2020)

\bibitem{gamba2024exit}
Gamba, D., Yu, Y., Yuan, Y., Schoenebeck, G., Romero, D.M.: Exit ripple effects: Understanding the disruption of socialization networks following employee departures. In: WWW. pp. 211--222 (2024)

\bibitem{yang23HGNAS}
Gao, Y., Zhang, P., Zhou, C., Yang, H., Li, Z., Hu, Y., Yu, P.S.: Hgnas++: Efficient architecture search for heterogeneous graph neural networks. TKDE  \textbf{35}(9),  9448--9461 (2023)

\bibitem{grover2016node2vec}
Grover, A., Leskovec, J.: node2vec: Scalable feature learning for networks. In: SIGKDD. pp. 855--864 (2016)

\bibitem{hamilton2017inductive}
Hamilton, W., Ying, Z., Leskovec, J.: Inductive representation learning on large graphs. NIPS  \textbf{30} (2017)

\bibitem{hu2020heterogeneous}
Hu, Z., Dong, Y., Wang, K., Sun, Y.: Heterogeneous graph transformer. In: WWW. pp. 2704--2710 (2020)

\bibitem{jia2023enhancing}
Jia, Y., Zou, D., Wang, H., Jin, H.: Enhancing node-level adversarial defenses by lipschitz regularization of graph neural networks. In: SIGKDD. pp. 951--963 (2023)

\bibitem{jin2023heterformer}
Jin, B., Zhang, Y., Zhu, Q., Han, J.: Heterformer: Transformer-based deep node representation learning on heterogeneous text-rich networks. In: SIGKDD. pp. 1020--1031 (2023)

\bibitem{jin2023transferable}
Jin, Y., Chen, K., Yang, Q.: Transferable graph structure learning for graph-based traffic forecasting across cities. In: SIGKDD. pp. 1032--1043 (2023)

\bibitem{kipf2016semi}
Kipf, T.N., Welling, M.: Semi-supervised classification with graph convolutional networks. arXiv preprint arXiv:1609.02907  (2016)

\bibitem{li2023metapath}
Li, M., Cai, X., Xu, S., Ji, H.: Metapath-aggregated heterogeneous graph neural network for drug--target interaction prediction. Briefings in Bioinformatics  \textbf{24}(1),  bbac578 (2023)

\bibitem{liang2022meta}
Liang, X., Ma, Y., Cheng, G., Fan, C., Yang, Y., Liu, Z.: Meta-path-based heterogeneous graph neural networks in academic network. International Journal of Machine Learning and Cybernetics pp. 1--17 (2022)

\bibitem{liu2023knowledge}
Liu, L., Tong, H.: Knowledge graph reasoning and its applications. In: SIGKDD. pp. 5813--5814 (2023)

\bibitem{liu2023generative}
Liu, S., Cai, Q., He, Z., Sun, B., McAuley, J., Zheng, D., Jiang, P., Gai, K.: Generative flow network for listwise recommendation. In: SIGKDD. pp. 1524--1534 (2023)

\bibitem{liu2023rhgn}
Liu, X., Zhang, K., Liu, Y., Chen, E., Huang, Z., Yue, L., Yan, J.: Rhgn: Relation-gated heterogeneous graph network for entity alignment in knowledge graphs. In: ACL. pp. 8683--8696 (2023)

\bibitem{lv2021we}
Lv, Q., Ding, M., Liu, Q., Chen, Y., Feng, W., He, S., Zhou, C., Jiang, J., Dong, Y., Tang, J.: Are we really making much progress? revisiting, benchmarking and refining heterogeneous graph neural networks. In: SIGKDD. pp. 1150--1160 (2021)

\bibitem{MaYLMC24HetGPT}
Ma, Y., Yan, N., Li, J., Mortazavi, M.S., Chawla, N.V.: Hetgpt: Harnessing the power of prompt tuning in pre-trained heterogeneous graph neural networks. In: WWW. pp. 1015--1023. {ACM} (2024)

\bibitem{mao2024hetfs}
Mao, X., Chen, Z., He, Z., Jing, Y., Zhang, K., Wang, X.S.: Hetfs: a method for fast similarity search with ad-hoc meta-paths on heterogeneous information networks. WWW  \textbf{27}(6), ~66 (2024)

\bibitem{mikolov2013distributed}
Mikolov, T., Sutskever, I., Chen, K., Corrado, G.S., Dean, J.: Distributed representations of words and phrases and their compositionality. NIPS  \textbf{26} (2013)

\bibitem{muppasani2024expressive}
Muppasani, B., Narayanan, V., Srivastava, B., Huhns, M.N.: Expressive and flexible simulation of information spread strategies in social networks using planning. In: AAAI. vol.~38, pp. 23820--23822 (2024)

\bibitem{perozzi2014deepwalk}
Perozzi, B., Al-Rfou, R., Skiena, S.: Deepwalk: Online learning of social representations. In: SIGKDD. pp. 701--710 (2014)

\bibitem{schlichtkrull2018modeling}
Schlichtkrull, M., Kipf, T.N., Bloem, P., Van Den~Berg, R., Titov, I., Welling, M.: Modeling relational data with graph convolutional networks. In: ESWC. pp. 593--607. Springer (2018)

\bibitem{SHAN24KPI-HGNN}
Shan, D., Du, X., Wang, W., Wang, N., Liu, A.: Kpi-hgnn: Key provenance identification based on a heterogeneous graph neural network for big data access control. Information Sciences  \textbf{659},  120059 (2024)

\bibitem{Shi2022hgnn}
Shi, C.: Heterogeneous Graph Neural Networks, pp. 351--369. Springer Nature Singapore, Singapore (2022)

\bibitem{vaswani2017attention}
Vaswani, A., Shazeer, N., Parmar, N., Uszkoreit, J., Jones, L., Gomez, A.N., Kaiser, {\L}., Polosukhin, I.: Attention is all you need. NIPS  \textbf{30} (2017)

\bibitem{velickovic2017graph}
Velickovic, P., Cucurull, G., Casanova, A., Romero, A., Lio, P., Bengio, Y., et~al.: Graph attention networks. stat  \textbf{1050}(20),  10--48550 (2017)

\bibitem{wang2019heterogeneous}
Wang, X., Ji, H., Shi, C., Wang, B., Ye, Y., Cui, P., Yu, P.S.: Heterogeneous graph attention network. In: WWW. pp. 2022--2032 (2019)

\bibitem{WU23Heterogeneous}
Wu, Y., Fu, Y., Xu, J., Yin, H., Zhou, Q., Liu, D.: Heterogeneous question answering community detection based on graph neural network. Information Sciences  \textbf{621},  652--671 (2023)

\bibitem{yan2021relation}
Yan, Q., Zhang, Y., Liu, Q., Wu, S., Wang, L.: Relation-aware heterogeneous graph for user profiling. In: ICKM. pp. 3573--3577 (2021)

\bibitem{yang2024fine}
Yang, M., Zhu, M., Wang, Y., Chen, L., Zhao, Y., Wang, X., Han, B., Zheng, X., Yin, J.: Fine-tuning large language model based explainable recommendation with explainable quality reward. In: AAAI. vol.~38, pp. 9250--9259 (2024)

\bibitem{Yang23Simple}
Yang, X., Yan, M., Pan, S., Ye, X., Fan, D.: Simple and efficient heterogeneous graph neural network. In: AAAI. pp. 10816--10824. {AAAI} Press (2023)

\bibitem{yu23RHGNN}
Yu, L., Sun, L., Du, B., Liu, C., Lv, W., Xiong, H.: Heterogeneous graph representation learning with relation awareness. TKDE  \textbf{35} (2023)

\bibitem{Yu23KGTrust}
Yu, Z., Jin, D., Huo, C., Wang, Z., Liu, X., Qi, H., Wu, J., Wu, L.: Kgtrust: Evaluating trustworthiness of siot via knowledge enhanced graph neural networks. In: WWW. pp. 727--736. {ACM} (2023)

\bibitem{zang2023commonsense}
Zang, X., Hu, B., Chu, J., Zhang, Z., Zhang, G., Zhou, J., Zhong, W.: Commonsense knowledge graph towards super app and its applications in alipay. In: SIGKDD. pp. 5509--5519 (2023)

\bibitem{zhang2019heterogeneous}
Zhang, C., Song, D., Huang, C., Swami, A., Chawla, N.V.: Heterogeneous graph neural network. In: SIGKDD. pp. 793--803 (2019)

\bibitem{zhang23pagelink}
Zhang, S., Zhang, J., Song, X., Adeshina, S., Zheng, D., Faloutsos, C., Sun, Y.: Page-link: Path-based graph neural network explanation for heterogeneous link prediction. In: WWW. p. 3784–3793. WWW '23, Association for Computing Machinery (2023)

\bibitem{zhong2023knowledge}
Zhong, Z., Mottin, D.: Knowledge-augmented graph machine learning for drug discovery: From precision to interpretability. In: SIGKDD. pp. 5841--5842 (2023)

\bibitem{zhu2023hagnn}
Zhu, G., Zhu, Z., Chen, H., Yuan, C., Huang, Y.: Hagnn: Hybrid aggregation for heterogeneous graph neural networks. arXiv preprint arXiv:2307.01636  (2023)

\bibitem{zhu23AutoAC}
Zhu, G., Zhu, Z., Wang, W., Xu, Z., Yuan, C., Huang, Y.: Autoac: Towards automated attribute completion for heterogeneous graph neural network. In: ICDE. pp. 2808--2821 (2023)

\end{thebibliography}




\end{document}